\documentclass[10pt,twocolumn,letterpaper]{article}

\usepackage{cvpr}
\usepackage{times}
\usepackage{epsfig}
\usepackage{graphicx}
\usepackage{amsmath}
\usepackage{amssymb}
\usepackage{amscd,amsbsy,latexsym,url,bm,amsthm,listings,xcolor,float,color,subcaption,makecell}

\usepackage[breaklinks=true,bookmarks=false]{hyperref}

\cvprfinalcopy 

\def\cvprPaperID{****} 

\begin{document}

\title{Secrets in Computing Optical Flow by Convolutional Networks}

\author{
Junxuan Li\\
The Australian National University\\
{\tt\small u5990546@anu.edu.au}
}

\maketitle

\begin{abstract}
Convolutional neural networks (CNNs) have been widely used over many areas in compute vision. Especially in classification. Recently, FlowNet and several works on optical estimation using CNNs shows the potential ability of CNNs in doing per-pixel regression. We proposed several CNNs network architectures  that can estimate optical flow, and fully unveiled the intrinsic different between these structures.
\end{abstract}
\section{Introduction}
That motion shown in two adjacent images or frames in an video is called optical flow. Optical flow estimation is the algorithms to estimating motion between two images or frames.  The most general version of optical flow is to compute the independent of motion at each pixel of an image. 

As recently, the convolutional neural networks have been widely used in many areas of computer vision.  We want estimate the optical flow by introducing a network architecture without any full connective layers and downsampling. Because we have a intuition that if we want to get a output that is as the same size as input, the downsampling of a network maybe unnecessary. However, as the experiments went on, we found that this kind of network performs bad no matter what kind of techniques were used in training until the downsampling layers were introduced. Several experiments and researches were done in order  to gave a reasonable explanation and unveil the intrinsic problems inside   this phenomenon.

Finally, we proposed a deeper network with downsampling and upsampling layers that works quite well in this problem. This new network will take two resources as input. One is the raw images pairs, the other is an approximately flow that derive from some other quick methods. We want this CNNs can both fine tuning the approximately flow and take it as guide to compute a more accurately flow.

\section{Related work}
\textbf{Classical models:} Optical flow has been analysis well over past 30 years. The most widely used models that defines the problem in optical flow is the brightness constancy and spatial smoothness assumption introduce by Horn and Schunck \cite{horn1981determining}. Many other method make some extract assumption base on these two models. Such as classical++ \cite{sun2010secrets} apply the median filter together with this model to eliminate the outliers in the quadratic formulation. EpicFlow \cite{revaud2015epicflow} introduce edge-preserving matches so that this model would be more robust to large displacement. However, all the method base on this model can not deal with all the constraint of optical flow in realistic scenes. Such as brightness change that are conducted by changing of light field rather than the motion of the object. These methods are all base on constraints and model that design by human, not any learning from realistic dataset. The progress of these kinds of method will demand more and more human design patterns as long as people want these models to fit the realistic better. It seems that a automatic learning way to do these pattern design is valuable which may highly decrease the human design patterns among the estimation. Also, these methods are usually slow.

\textbf{CNNs}: Now the CNNs is a new technology that performs well on most of the areas in computer vision such as classification \cite{lecun1989backpropagation} \cite{krizhevsky2012imagenet} and recognition \cite{hijazi2015using}. People found out that the convolution network is very suitable in many computer vision areas, for it can extract high-level feature maps from each images over large spaces. Especially as the growth of computability in GPU and size of dataset we can get from Internet, learning a deep complex neural network become more and more computation feasible. But nowadays, most of the task that CNNs do is classification. It also shown its ability in per-pixel regression such as semantic segmentation \cite{long2015fully} and depth estimation \cite{eigen2014depth}. We want to discover ability that CNNs have in per-pixel regression tasks.

\textbf{CNNs on optical flow:} Recently, there are many works that introduce some different CNNs in doing  optical flow computation. FlowNet \cite{dosovitskiy2015flownet} is the first one trying CNNs on optical flow. And its network architecture is widely used by other researches \cite{ilg2016flownet} \cite{ranjan2016optical}. This paper not only provide a thought in how to build a network for optical flow, but also generated a big dataset that can be used in training optical flow, the FlyingChairs. But the shortage among this paper is also obvious. As the FlyingChairs is generated by human and its structure of motion is rather simple, only contains rigid motion. So the performance of this network is highly restricted by training dataset, as it will overfitting on training data. It can be seen that on it performs far more better on testing data in FlyingChairs. But not comparable with many other start-of-art methods in other dataset. 

\textbf{Unsupervised CNNs:} Many method use unsupervised CNNs \cite{jason2016back} \cite{ahmadi2016unsupervised} \cite{zhu2017guided}. \cite{zhu2017guided} only change the ground truth in FlowNet to be the results of state-of-the-art classical estimators. So that they can apply this learning to some realistic dataset and not need to be restricted by the ground truth of a dataset. But this method  highly relies on the accuracy of state of art method. \cite{jason2016back} using brightness
constancy and motion smoothness as the loss function of the network rather than a ground truth flow. But the performance is also been restricted by the goodness of the model and assumption. It can not cover all the situations in real life. But these two unsupervised learning indeed get rid of the constraint of dataset. As we can get more data from real life rather than random generate from a single data. It can prevent overfitting to a special situation.

\section{First trial on Network architectures}
In our opinions, the network \cite{krizhevsky2012imagenet} would like to use layers with downsampling, by which they can shrink the features over whole image, in order to use a softmax or full connective layer to do the classification. But if our tasks is to do a pixel-to-pixel estimation, the downsampling layer maybe redundant, since the most of them \cite{dosovitskiy2015flownet} \cite{zhu2017guided} using downsampling layers have to recover the original resolution by several upconvolution layers. 

\subsection{Network without downsampling}
\begin{figure}
	\centering
	\includegraphics[width=\linewidth]{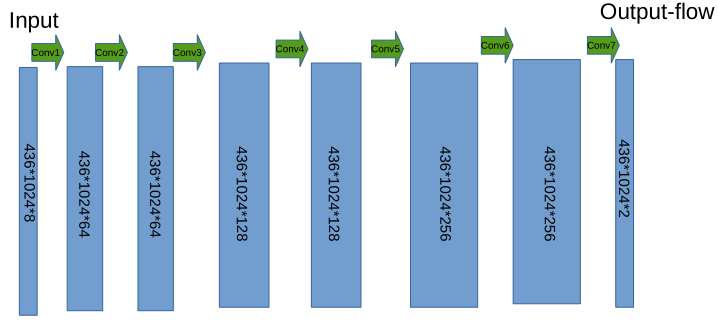}
	\caption{A small network without any downsampling. The green arrow with text \texttt{Conv} is Convolutional block with 5 by 5 filter size. One convolutional block will include a convolution and batch normalization and RELU. The number in each block is the input size of each block. e.g. $436\times 1024\times 8$ is the input size of first block.}
	\label{fig:smallflownet}
\end{figure}

So we proposed a network without downsampling. It is designed as shown in Figure \ref{fig:smallflownet}. For each convolution block it has, it only contains convolution with 5 by 5 filter window size, batch normalization and RELU, all with 1 pixel stride. For each layers, the input of it is labelled in the figure. Typically, the input of first layer is $H\times W\times 8$ where  $H,W$ is the height and width of original images. As the limited of computation of GPU, we could not make this network any bigger. To train this network, we need to store at least  $H\times W\times 906$ (combine the input of each layer) values in forward and backward of the network. But only with 22k parameters (906 convolution filters with 5 by 5 windows) to train. So in total, this network should need less data to train, as the parameters is less, and faster to test, as the layer is shallow.

\subsection{Error function}
As training loss we use the endpoint error (EPE) \cite{baker2011database}, which is the standard error measure for
optical flow estimation.

The normalised endpoint error (NE) between an estimation flow $I(x,y)$ and a ground-truth flow $I_{GT}(x,y)$ is given by:
\begin{align}
NE = \left[ \sum_{(x,y)}^{} \dfrac{(I(x,y)-I_{GT}(x,y))^2}{\|\nabla I_{GT}(x,y)\|^{2}+ \epsilon}\right]^{\frac{1}{2}}
\end{align}

For CNNs is using back-propagation to update the parameter. We need to derive its partial derivative over each pixel $I(x_i,y_i)$:
\begin{align}
\frac{I(x_i,y_i)-I_{GT}(x,y) }{[ (\|\nabla I_{GT}(x,y)\|^{2}+ \epsilon)\sum_{(x,y)}^{} (I(x,y)-I_{GT}(x,y))^2]^{\frac{1}{2}}}
\end{align}

\subsection{Intrinsic problem in this network}
However, the performance on this network is bad. The comparison of its results will be shown in Results section. In general, we will easily overfit this network so that it will perform quite well  on the training dataset while bad on the testing dataset. Besides, the output of this network seems to be very sensitive to the colour-map of input. But as we know that optical flow should be derived based on relative displacements between frames rather than colour. 

So we can conclude that this network can only learning some basic features of an images, such as edges, colour and brightness. It can not extract the high-level features that can be  used to compute optical flow. What kind of problem will rise situation like this? First of all, the training dataset should be enough, as we only have few parameters to train. Also, the performance in the validation dataset show us that it is never getting better till the end of training. 

Can the network get any deeper, so that we can extract and process more features than before? No, we are not using any downsampling here, so the memory demanding of this network will increase rapidly after 4 convolution layers and 6 convolution layer is the most layer that we could try temporary in a 11GB GPU. Even if we can get a bigger GPU to train it, the time and cost to train this big network is out of our consideration. 

According to Zeiler \etal \cite{zeiler2014visualizing}, they explored how discriminative the features are in different layers.  As the feature hierarchies become deeper, a better performance they can get using a linear SVM to do the classification problem. This shows that neural networks can learn increasing powerful features over layers. They also build a way to reconstruct the high-level features stored in low-dimensional space by an operation called unpooling. It seems that as the layer being deeper, the features it generate is more specific and discriminative. Usually the first 3 layers are edge and colour features which are very basic and not informative. But as the pooling and convolution get this network deeper and smaller. The features that been extracted show high discriminative of an object and even if the resolution of a feature is small, they can also construct the intact information in original images. These unpooling and deconvolution techniques are also been used in several papers doing semantic segmentation \cite{long2015fully} \cite{noh2015learning} which is also a task doing pixel-to-pixel estimation. By doing this, they can extract high-level features from original images over and large areas by  pooling or downsampling layers. An output of  original resolution can also been reconstruct by concat information from high-resolution with informative but low-resolution features. Been inspired by all the meaningful of doing downsampling, we proposed and new  network architecture that can both make used of high-level features and reduce the computation while upsampling.

\section{Final Network architecture}
\begin{figure*}
	\centering
	\includegraphics[width=\textwidth]{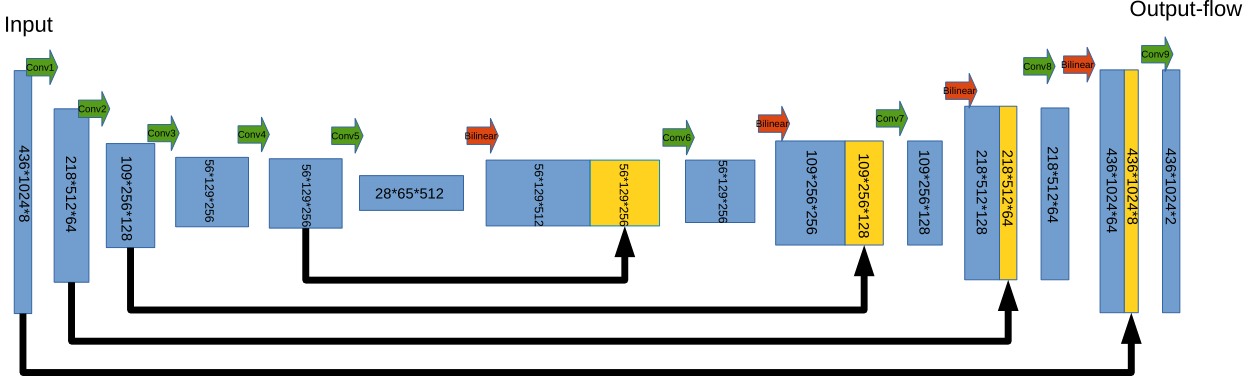}
	\caption{Final version of our network. The green arrow with text \texttt{Conv} is Convolutional block, it only contains pooling when it need to do the downsampling.  The third dimension of feature maps increases as the network go deeper, it is doubling after each convolutional block with max pooling stride two. The red arrow with text \texttt{Bilinear} is the bilinear upsampling block. It will simply bilinear resize the input to a twice big space. The black arrow from previously layers to post layers is the concat action which simply concatenate two layer in third dimension.}
	\label{fig:flownet}
\end{figure*}

We design an network shown in Figure \ref{fig:flownet}. All convolutions in our network is using a 1 pixel stride and all pooling is with 2 pixels stride, all with appropriate padding. When a convolution block is apply to a layer, it will do a series action, first convolution, then batch normalization, then RELU, and may contains a pooling operation, which depends on whether or not we will do a downsampling in this layer. As we do downsampling over the layers, the output is thus at half the resolution of input resolution. After four convolution blocks with downsampling, we will get $1/16$ of the original size of input. Apparently, this kind of size could not be output as a dense optical flow. So we design a bilinear upsampling layer which can  to obtain an output twice big as the input. A flow field of original resolution can be obtain after several bilinear upsampling layer is applied. 

\subsection{Network input}
The input of the network is a  $H\times W\times 8$ space of pixels made by combining two adjacent RGB images and its approximate flow. First six channels of input is the RGB value of two frames, the other two channels in the end is the uv map of approximate flow. The reason we also want to take an approximate flow as input is that if we only train this network with RGB channel, the network can not learn much from the dataset. The performance is not as good as a approximate flow. So we want to add this to guide the network go for a better training. And this approximate flow we add indeed improve the performance of the network, so that we can make this it work even with a small training dataset.

This approximate flow is only used as a guided flow so that network can estimation the final flow easily, it can highly decrease the demanding of training data while not make much influence of flow accuracy. 
For this approximate flow is just a pre-computation of the input, it should not be  time consuming. In fact, we want this pre-computation as fast as possible while not losing the general information of a coarse optical flow. Because one of the advantages of CNNs is the speed on running time. We want to make the this optical flow estimate outperform other classical methods in run times. 

There are several methods that perform well in optical estimation such as EpicFlow \cite{revaud2015epicflow} and LDOF \cite{brox2009large}. But both of them take more than 10s to compute a frame. DIS-Fast \cite{kroeger2016fast} and PCA-Flow \cite{wulff2015efficient} is much faster than previously two, they take around $0.1$ seconds to do the estimation. These two could be a good method to do the approximate flow computation. However the codes to implement these two method is not released. I try to use Block matching \cite{barjatya2004block}, a patch based method which could do a coarse motion estimation.

\subsection{Bilinear upsampling}
Applying pooling layers or convolutions with stride two will make CNNs good at extracting high-level features that human unreadable from images.  Besides, downsampling is necessary to make network training computationally feasible. The networks without any pooling or convolution with 2 pixel stride will not only demands a huge computation ability of computer, but also can not aggregate the interesting features over large areas of big feature maps. However, pooling will cause output layer have a half resolution of input feature maps, so in order to compute dense flow in each pixel, we need a refinement in those sparse feature maps to get results of original resolution.

In 'FlowNet Simple' by Dosovitskiy \etal \cite{dosovitskiy2015flownet}, they used a  'upconvolutional' layers, consisting of unpooling (extending the feature maps, as opposed to pooling) and a convolution. While we used a computationally less expensive bilinear upsampling. To perform the upsampling and continue with feature dimensional shrinking, we apply the 'Bilinear upsampling' to a small feature maps, and concatenate its output with suitable size of previous feature maps in the network and an apply convolution block after concatenation. By doing this we  can preserve  the dense  information passed from previous feature maps together with the high level information over large areas that extracted by pooling operations. Each a bilinear upsampling and a convolution will increase the resolution twice and shrink the third dimension of a feature maps to one third. We repeat it 4 times. At the last convolution block, we make its output channel 2 to be a prediction layer, resulting in a dense optical flow for which the resolution is the same as original images.

\section{Implementation}

\subsection{Training dataset}
Unlike traditional approaches such as \cite{chen2013large} only learning a few parameters, neural networks have million parameters to train from scratch. Besides,  it is also different from some problems in images classification, the dense ground truth optical flow of two images is hard to get.  The following dataset is suitable for training and testing.

\textbf{MPI-Sintel} dataset \cite{butler2012naturalistic} is a large dataset available which contains $1041$ training image pairs. It displacement of ground truth flow vary from small magnitude to large, which will make the training go overfitting. \textbf{Middlebury} dataset \cite{baker2011database} has $8$ image pairs with ground truth. Its magnitude of flow  usually less than $9$ pixels. We use this dataset as the testing dataset.

\subsection{Training CNNs}
For training CNNs, we use \texttt{'MatConvNet'} \cite{vedaldi15matconvnet} framework with few changes. We choose standard stochastic gradient descent with momentum as optimisation method. We use a small mini-batches of $6$ image pairs.  To prevent from overfitting during training, we split the Sintel dataset into 900 training and 141 validation pairs.
We start with learning rate $\lambda = 10^{-7}$, and reduce to $\lambda = 10^{-8}$ after half iterations.

\section{Results}
\begin{figure}
	\centering
	\includegraphics[width=0.5\linewidth]{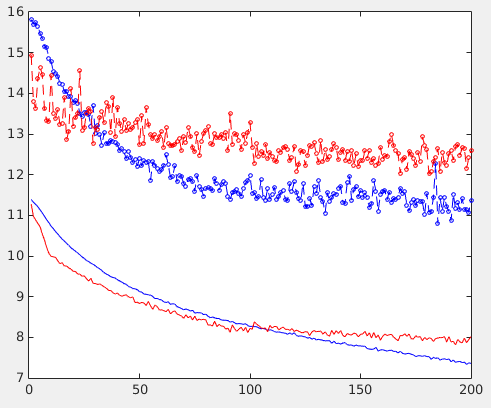}
	\caption{Training curve, blue line is the error of FinalNet, red line is the error of PlainNet. Solid line below is represent for training dataset error, dashdot line is represent for validation dataset error. X coordinate is number of epochs, Y coordinate is endpoint error.}
	\label{fig:curve}
\end{figure}

The training curve of first trial network (denoted as PlainNet) and final network(denoted as FinalNet) is shown in Figure \ref{fig:curve}. We can see that error of validation on PlainNet does not change much, but the error in training  dataset keep reducing, which means that PlainNet learning few things from intrinsic logic of optical. While FinalNet can keep both validation and training reduce until 100 epochs. The FinalNet seems to converge in around 100 epochs, and start a little bit overfitting from 100 to 200 epochs. For the error of   training dataset is keep reducing while the error of  validation dataset  is not changed. So we choose network in epoch 150 to be the final parameters of network. 

It can be seen from Table \ref{tab:epelos} that, for the data that we never met during the training (Sintel Validation and Middlebury Testing) we can get around $8\%$ improvement on FinalNet. But the PlainNet is only working on training dataset, which still shown the overfitting of it.

\begin{figure*}
\centering
\begin{subfigure}{1\textwidth}
	\centering
	\includegraphics[width=0.19\textwidth]{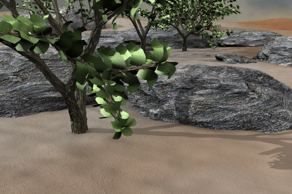}
	\includegraphics[width=0.19\textwidth]{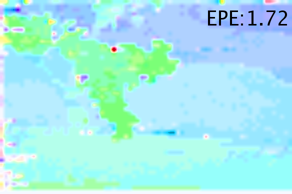}
	\includegraphics[width=0.19\textwidth]{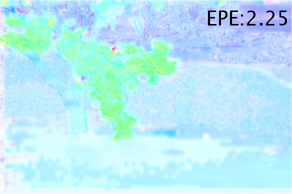}
	\includegraphics[width=0.19\textwidth]{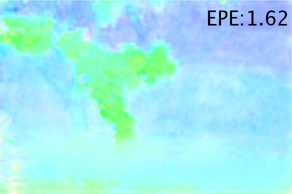}
	\includegraphics[width=0.19\textwidth]{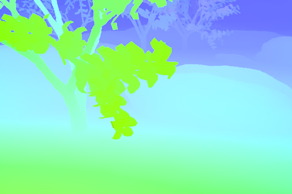}
\end{subfigure}

\begin{subfigure}{1\textwidth}
	\centering
	\includegraphics[width=0.19\textwidth]{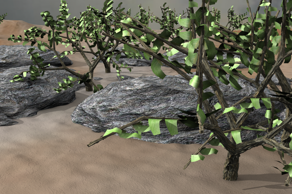}
	\includegraphics[width=0.19\textwidth]{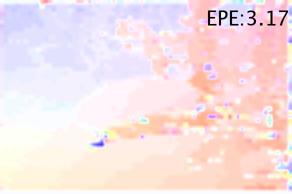}
	\includegraphics[width=0.19\textwidth]{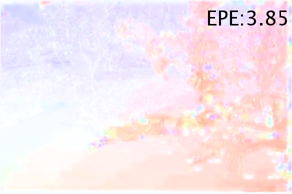}
	\includegraphics[width=0.19\textwidth]{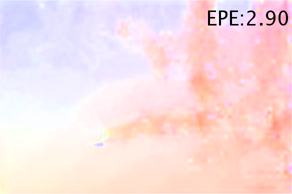}
	\includegraphics[width=0.19\textwidth]{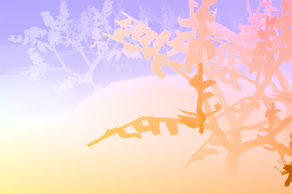}
\end{subfigure}
\begin{subfigure}{1\textwidth}
	\centering
	\includegraphics[width=0.19\textwidth]{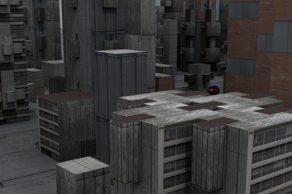}
	\includegraphics[width=0.19\textwidth]{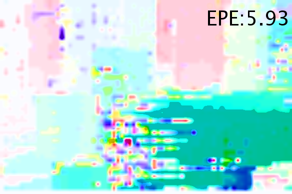}
	\includegraphics[width=0.19\textwidth]{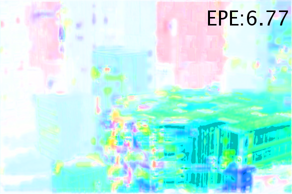}
	\includegraphics[width=0.19\textwidth]{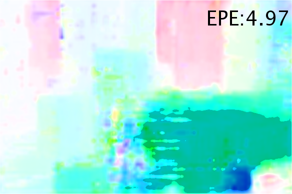}
	\includegraphics[width=0.19\textwidth]{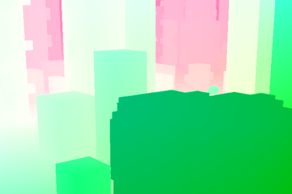}
\end{subfigure}
\begin{subfigure}{1\textwidth}
	\centering
	\includegraphics[width=0.19\textwidth]{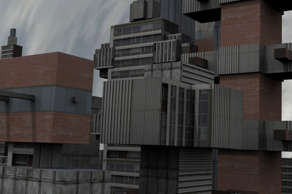}
	\includegraphics[width=0.19\textwidth]{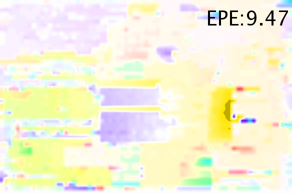}
	\includegraphics[width=0.19\textwidth]{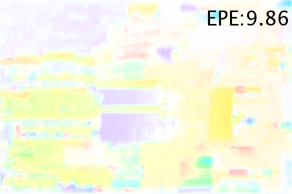}
	\includegraphics[width=0.19\textwidth]{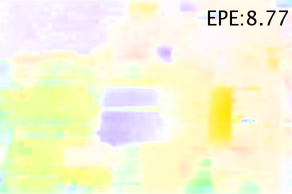}
	\includegraphics[width=0.19\textwidth]{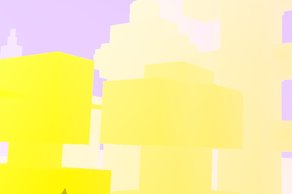}
\end{subfigure}
\begin{subfigure}{1\textwidth}
	\centering
	\includegraphics[width=0.19\textwidth]{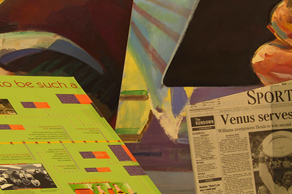}
	\includegraphics[width=0.19\textwidth]{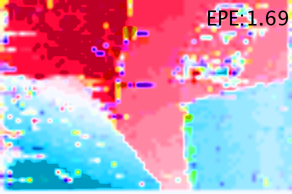}
	\includegraphics[width=0.19\textwidth]{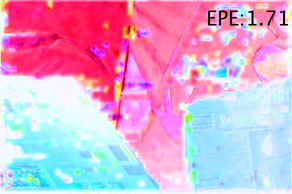}
	\includegraphics[width=0.19\textwidth]{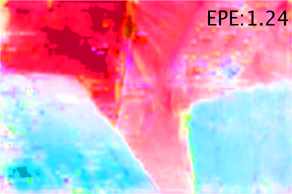}
	\includegraphics[width=0.19\textwidth]{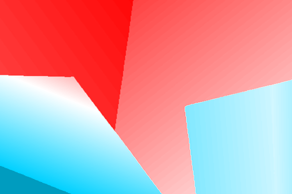}

\end{subfigure}
\caption{Examples of optical flow prediction on the Middlebury dataset by our network. In each row left to right: original image, block matching results, output of PlainNet,  output of FinalNet, ground truth flow. Endpoint error is shown for every frame. Note that even though the EPE of block matching results is usually worse than that of output of network, the networks can refine the flow and make it more accurate.}
\label{fig:middresult}
\end{figure*}

There are some  example results shown in Figure \ref{fig:middresult}. We can see from the images that most of the outliers in block matching results can be eliminate in output flow by FinalNet. And it can also help to correct some deviation that generate in block matching. While PlainNet remains the edges and colour features of the original images, it can not predict flow.

In Table \ref{tab:time} we show the runtime of each image pair by different methods. Although our method is not as good as others in accuracy, we can out perform most of the classical method in runtimes.  Besides, both the runtimes and performance of FinalNet  outperform PlainNet.
Both training and testing of the networks is been done by NVIDIA GTX 1080 GPU while the computation of approximately is on CPU. The  runtimes of other method is taken from their papers. 

\begin{table}
	\centering
	\begin{tabular}{ c | c | c  |c}
		\hline			
		EPE 				& \makecell{Block\\ Matching} 	 & PlainNet & FinalNet \\
		\hline
		Training 	& 9.19 				  	&	8.01 & 7.13\\
		Validation    & 12.14 			 	&	12.61	& 11.22\\
		Middlebury 			&  3.09 				& 3.61 & 2.90\\
		\hline  
	\end{tabular}
	\caption{ Average endpoint errors (in pixels) of my networks}
	\label{tab:epelos}
\end{table}

\begin{table}
	\centering
	\begin{tabular}{ c | c | c  }
		\hline			
		Method 		& \makecell{Time \\ CPU} 	& \makecell{Time \\ GPU}  \\
		\hline
		EpicFlow \cite{revaud2015epicflow} 		& 16 			& -  		\\
		LDOF Validation \cite{brox2009large}   	& 65 			& 2.5 			\\
		FlowNetS \cite{dosovitskiy2015flownet}	&		-	&	0.08	\\
		FlowNet2-s \cite{ilg2016flownet}		& - &   7 \\
		Our-PlainNet &  0.3 &   0.16\\
		Our-FinalNet &  0.3 &   0.04\\
		\hline  
	\end{tabular}
	\caption{Running time compare with other methods}
	\label{tab:time}
\end{table}

\section{Discussions}
The most important thing in this network is the downsampling. If we want to get a output that is as the same size as input, why do we want to do downsampling and upsampling in the network? Is all convolution without any stride two a good design? The answer is clearly no after we  tried an network without any downsampling.  For PlainNet will not shrink the features over layers, its computation demands much more than FinalNet. Besides, it seems that PlainNet will overfit much. The output of PlainNet is very sensitive to the color of a image rather than displacement. We think the reason would be that downsampling can help network to extract useful features. And as the network get deeper, it can prevent overfitting.

\section{Conclusions}
In summary, we show how important the downsampling layers are in a network. We also have proposed a new network that can do pixel-to-pixel estimation by combining high-level features  with images details. Our approach can also combine with some other state-of-art optical flow estimation methods in the future, and see how it work with them.

{\small
\bibliographystyle{ieee}
\bibliography{egbib_engn8535}
}

\end{document}